\documentclass{article}

\usepackage{PRIMEarxiv}

\usepackage[utf8]{inputenc}
\usepackage[T1]{fontenc}    
\usepackage[numbers,sort&compress]{natbib}
\usepackage{hyperref}       
\usepackage{url}           
\usepackage{doi}
\usepackage{booktabs}       
\usepackage{multirow}       
\usepackage{amsmath}
\usepackage{amsfonts}       
\usepackage{nicefrac}       
\usepackage{microtype}      
\usepackage{lipsum}
\usepackage{fancyhdr}       
\usepackage{graphicx}       
\usepackage{float}
\graphicspath{{media/}}     
\newcommand{\iconlink}[3]{%
  \href{#1}{%
    \raisebox{-0.15em}{\includegraphics[height=1em]{#2}}~#3%
  }%
}
\title {HC-RAG: Evidence-Centric Retrieval-Augmented Generation over Heterogeneous Financial Filings}

\author{
  Siyuan Chen\\
  Sun Yat-sen University\\
  Zhuhai\\
  \texttt{chensy363@mail2.sysu.edu.cn} \\
   \And
  Huaye Tan\\
  Central South University\\
  Changsha\\
  \texttt{harveytan215@gmail.com} \\
  \And
  You Li\\
  Sun Yat-sen University\\
  Zhuhai\\
  \texttt{liyou39@mail2.sysu.edu.cn} \\
   \And
  Jiajun Liang\\
  Sun Yat-sen University\\
  Zhuhai\\
  \texttt{liangjj85@mail2.sysu.edu.com} \\
}

\begin{document}
\maketitle

\begin{abstract}
Financial question answering over annual reports requires more than retrieving semantically similar passages. It often involves identifying relevant companies and fiscal years, locating standardized filing sections, collecting textual and tabular evidence, and checking answers against the original documents. Existing RAG systems, however, usually flatten long filings into unordered chunks, pay limited attention to the typed structure of financial reports, and use fixed text-table fusion strategies without considering query intent. To address these limitations, we propose \textbf{HC-RAG}, a hierarchical cross-modal retrieval-augmented generation framework for evidence-centric financial QA. HC-RAG organizes filings into a typed financial evidence graph with documents, sections, text units, table units, and metadata nodes. It retrieves evidence through document-section-unit paths, aligns textual and tabular evidence in a shared retrieval space, and routes evidence according to four semantic intents: calculation, trend, fact, and comparison. We further introduce \textbf{Multi-Doc-2025}, a benchmark containing 2,327 expert-verified QA pairs from 179 SEC 10-K filings of 87 S\&P 500 companies across fiscal years 2022--2024, with labels for intent, difficulty, and structural evidence attributes. Experiments on public financial QA benchmarks and Multi-Doc-2025 show that HC-RAG improves both answer quality and evidence localization, especially in long-document, table-related, and cross-document settings. HC-RAG outperforms RAPTOR by 6.6 F1 points on DocFinQA and GraphRAG by 10.9 F1 points on Multi-Doc-2025. Evidence-level analysis and ablation studies show that the improvements mainly come from more accurate section localization, table grounding, cross-document evidence aggregation, and intent-aware text-table routing.

\end{abstract}


\section{Introduction}
Currently, most financial institutions worldwide are designing their own AI agents. While intelligent customer service tools for the public are quite mature, professional analytical tools for analysts and practitioners are still relatively few. Financial statements are the most important analytical material for financial professionals, and intelligent agents and AI-enabled tools focused on financial statement analysis are constantly developing. Financial question answering, in particular, has significant practical and productive value. Financial Q\&A over annual reports depends on finding evidence that can be checked in the original filings. Unlike ordinary text documents, annual reports are usually long, structured, and contain both narrative text and financial tables. A Form 10-K filing often has tens of thousands of tokens and is organized into sections such as business overview, risk factors, MD\&A, financial statements, and notes. These sections are closely related to each other. For example, narrative explanations may refer to values in tables, risk disclosures may depend on accounting information, and many analyst-style questions require comparisons across companies or fiscal years. RAG is a useful way to make large language models answer questions based on external evidence~\cite{RAGNLP,RAGSurvey}, but financial filings impose a more structured retrieval problem than ordinary open-domain corpora because the supporting evidence is hierarchical, cross-document, cross-year, and text-table hybrid~\cite{Bloomberggpt,Financebench}.

Current RAG systems still have several problems in this setting. First, many methods split long filings into flat chunks and retrieve them as independent candidates. This breaks the original document-section-unit structure of annual reports and may also make important evidence harder to find in long documents, especially under the ``lost in the middle'' problem~\cite{LostinM2024}. Second, textual descriptions and financial tables provide different but related information. Text usually explains business context or accounting meaning, while tables provide exact numerical values. Simple table serialization or a single retrieval encoder may not align these two forms of evidence well~\cite{finqa2021,zhutatqa2021,TAPEX2022}. Third, different questions rely on different types of evidence. Calculation questions usually need exact table values, trend questions need temporal evidence, fact questions often depend on a specific statement, and comparison questions require evidence from different entities or periods. If the system always uses the same text-table fusion strategy, it may introduce irrelevant evidence instead of selecting the information needed by the query.

To address these issues, we propose \textbf{HC-RAG}, a hierarchical cross-modal retrieval-augmented generation framework for multi-document financial question answering. HC-RAG treats financial QA as an evidence retrieval problem over a heterogeneous financial evidence graph. In this graph, filings are represented with typed document, section, text-unit, and table-unit nodes. Retrieval is therefore not a global nearest-neighbor search over flat chunks, but a process of selecting evidence through valid graph paths. HC-RAG also aligns text and table representations with an asymmetric offline-online design, and uses query-aware feature routing based on four semantic intent classes: calculation, trend, fact, and comparison. We further introduce \textbf{Multi-Doc-2025}, a benchmark containing 2,327 QA pairs from 179 SEC 10-K filings of 87 S\&P 500 companies across fiscal years 2022--2024. The benchmark uses primary-company-disjoint splits and provides labels for cross-document, cross-year, hybrid-modal, difficulty-level, and intent-aware evaluation.

Our contributions can be summarized as follows:
\begin{itemize}
    \item We study financial QA over annual reports from the perspective of evidence retrieval. Instead of only focusing on whether the final answer is correct, we emphasize whether the system can locate verifiable evidence from structured financial filings.
    \item We propose HC-RAG, which uses a typed financial evidence graph, hierarchical evidence retrieval, and intent-aware text-table routing to support cross-document and cross-modal financial reasoning.
    \item We release \textbf{Multi-Doc-2025}, a benchmark built from SEC 10-K filings. It covers financial questions that involve different companies, fiscal years, and hybrid text-table evidence.
    \item We evaluate HC-RAG at both the answer level and the evidence level. The results show that better final answers are closely related to more accurate retrieval of documents, sections, tables, and cross-document evidence.
\end{itemize}

\section{Related Work}
\label{sec:headings}
\subsection{Long-Context and Graph-Augmented RAG}
RAPTOR, GraphRAG, and HippoRAG are representative hierarchical or graph-augmented RAG systems. These methods use tree or graph structures to help retrieve evidence across multiple steps~\cite{sarthi2024raptor,edge2024graphrag,gutierrezHippoRAGNeurobiologicallyInspired2025}. HippoRAG also shows that retrieval systems can borrow ideas from human memory mechanisms. HC-RAG follows a related idea, but focuses on the financial QA setting. Instead of modeling general memory, HC-RAG tries to model how analysts look for evidence in annual reports. It builds a typed evidence graph over filings, sections, text units, tables, companies, and fiscal years, so the retrieval process can better follow the structure of financial documents.

\subsection{Financial and Multimodal Document Question Answering}
Financial QA has developed from financial language understanding to tasks that require numerical reasoning, table-text understanding, and long-document evidence retrieval. FinBERT and later financial LLMs improve language modeling in the financial domain\cite{Finbert2019,cfgpt2023,FinGPT2023,instructfingpt2023,pixiu2023}. FinQA, ConvFinQA, TAT-QA, FinanceBench, and DocFinQA further show that financial QA needs arithmetic reasoning, factual grounding, and evidence localization in financial reports\cite{finqa2021,convfinqa2022,zhutatqa2021,Financebench,DocFinqa2025}. Research on document and table understanding has also introduced layout-aware models, multi-page document QA methods, table linearization methods, and tabular models such as TAPAS and TAPEX\cite{DocVQA2021,LayoutLMv32022,DocVQA2023,chartRAG2024,Chaintable2024,TAPAS2024,TAPEX2022}. These studies provide useful foundations for financial document QA. Most existing benchmarks and systems still focus on single documents, single tables, or local table-text contexts. In actual financial analysis, questions often involve comparisons across companies, tracking changes across fiscal years, and using narrative explanations together with financial statements. This motivates a benchmark and retrieval framework that can evaluate these requirements in the same setting.

\subsection{Adaptive Fusion and Query-Aware Routing}
Multimodal fusion studies how different information sources can be used together without weakening each other. Mixture-of-Experts architectures, adaptive gates, and unified multimodal encoders provide mechanisms for sparse activation and dynamic weighting\cite{SurveyRAG2024,Structure-aware2022,Unitab2022}. Many existing fusion methods mainly decide modality weights from the input representation or the retrieved context. For financial QA, the query itself should also play an important role. A question about calculating the current ratio usually depends more on table cells, while a question about liquidity risk usually depends more on explanatory text, even when both questions come from the same filing. HC-RAG uses calculation, trend analysis, fact finding, and comparison as query intent classes, and uses the predicted intent to route evidence between text and table candidates.

\section{Methodology}
\subsection{Analyst-Inspired Evidence Workflow}
HC-RAG is designed based on how financial analysts usually look for evidence in annual reports. When answering a question about filings, an analyst normally does not read the report as a set of unrelated passages. The analyst usually starts from the relevant company, fiscal year, and peer group, then moves to standard sections such as MD\&A, risk factors, financial statements, and notes. After that, the analyst collects the needed numerical values or narrative explanations and checks whether these pieces of evidence can support the final answer.

This process leads to several requirements for retrieval. The system should keep the hierarchy of annual reports, since document-level and section-level information can help narrow down the search before retrieving fine-grained evidence units. It should also distinguish different types of evidence, because textual descriptions, tables, company metadata, fiscal years, and financial metrics play different roles in financial QA. The retrieval process also needs to consider the intent of the question. Calculation, trend, fact, and comparison questions usually depend on different evidence patterns, and they should not use the same text-table fusion strategy in all cases.

HC-RAG implements these ideas through a financial evidence graph and an intent-aware evidence routing module. The evidence graph keeps the hierarchy of filings and the relations across documents. The routing module selects and combines textual and tabular evidence according to the semantic intent of the query.

\subsection{Problem Formulation and Evidence Graph}
\begin{figure*}[t]
    \centering
    \includegraphics[width=0.95\textwidth]{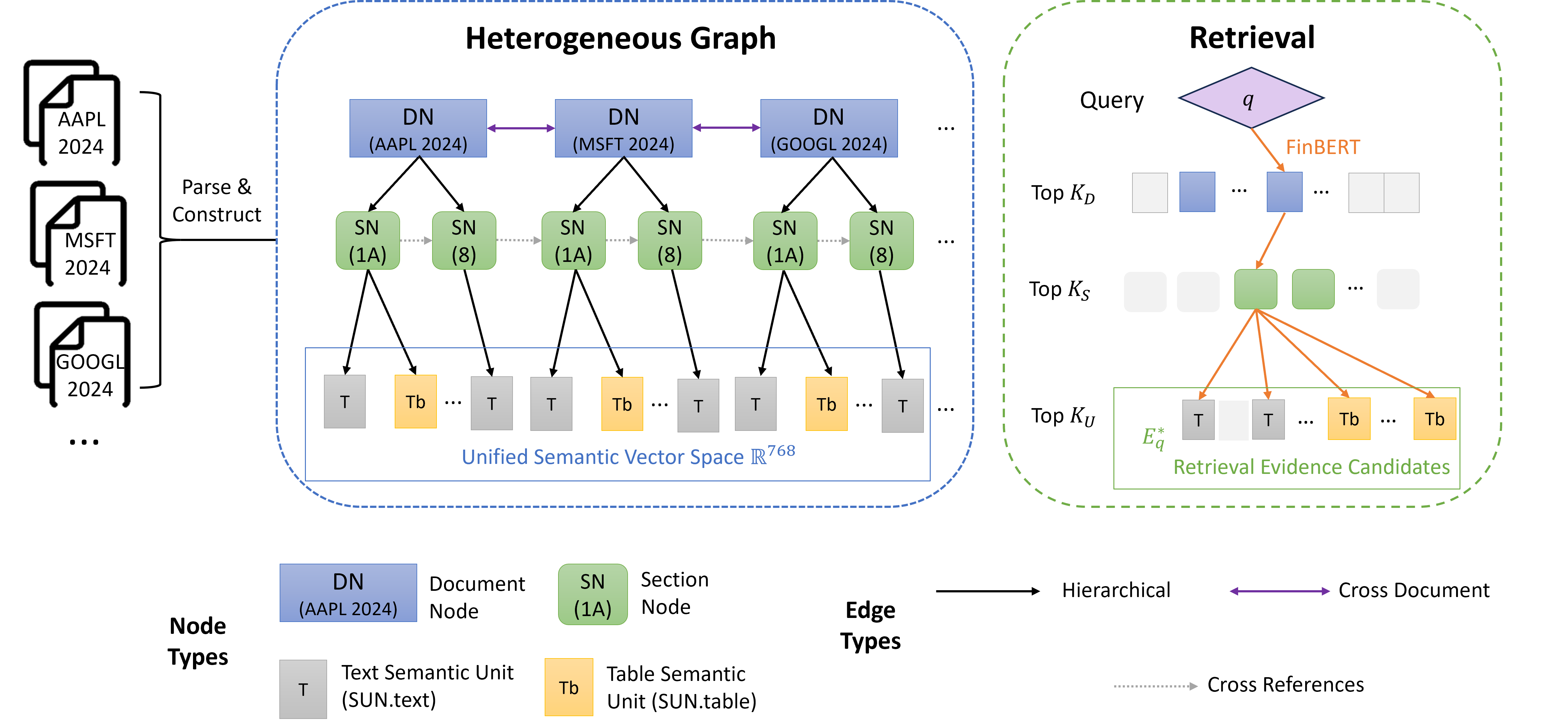}
    \caption{The indexing and retrieval framework of HC-RAG. The indexing layer builds a financial evidence graph from annual reports, where the table and text representations are mapped into a unified semantic vector space. The retrieval layer finds evidence through document-section-unit paths. }
    \label{fig:framework}
\end{figure*}

HC-RAG organizes the filing corpus into a financial evidence graph, as in the first blue box of Figure \ref{fig:framework}. The graph is written in a simple form as

\begin{equation}
\mathcal{G}=(\mathcal{V},\mathcal{E},\mathbf{X}),
\end{equation}

where $\mathcal{V}$ denotes all nodes, $\mathcal{E}$ denotes the relations between nodes, and $\mathbf{X}$ stores the text, table content, and metadata of each node. In this paper, the nodes mainly include document nodes, section nodes, text-unit nodes, table-unit nodes, and metadata nodes. Document nodes correspond to annual reports. Section nodes correspond to standard filing sections such as MD\&A, risk factors, financial statements, and notes. Text-unit nodes store narrative evidence, table-unit nodes store tabular evidence, and metadata nodes store information such as company, fiscal year, sector, and financial metric.

The edges in the graph are used to describe the basic relations inside and across filings. Containment edges connect documents with sections and connect sections with evidence units. Sequential edges keep the original order of sections and evidence units in the filing. Cross edges connect related companies, fiscal years, sectors, or financial metrics. Modal edges connect narrative descriptions with related table evidence. In this way, the graph keeps the reading structure of financial reports, from company and year to filing, from filing to section, and from section to textual or tabular evidence. For a filing-based question, a simple evidence path can be written as:

\begin{equation}
P: v_D \rightarrow v_S \rightarrow v_U,
\end{equation}

where $v_D$ is a document node, $v_S$ is a section node, and $v_U$ is a text or table evidence-unit node. This path means that the system first locates the relevant filing, then finds the related section, and finally retrieves the evidence unit that may support the answer. Cross-document and cross-year edges allow the system to move from one filing to related filings when the question involves comparison or trend analysis. Therefore, HC-RAG does not retrieve evidence from all chunks in a flat way. It retrieves evidence along the structure of financial reports, which makes the retrieved evidence easier to trace back to the source document. This design is related to graph-augmented retrieval systems such as RAPTOR, GraphRAG, and HippoRAG~\cite{sarthi2024raptor,edge2024graphrag,gutierrezHippoRAGNeurobiologicallyInspired2025}. The main difference is that HC-RAG builds the graph according to the structure of financial filings, rather than only relying on semantic similarity or LLM-generated relations.

\subsection{Hierarchical Graph Retrieval}
HC-RAG retrieves evidence in three steps: document retrieval, section retrieval, and evidence-unit retrieval. This process follows the way annual reports are usually read. The system first narrows the search to several relevant filings, then selects useful sections, and finally retrieves detailed text or table evidence from these sections. At the document level, HC-RAG ranks filing nodes by combining semantic similarity and metadata matching:

\begin{equation}
s_D(q,v_D)=\mathrm{sim}(q,v_D)+\alpha_D \mathrm{match}(q,v_D),
\end{equation}

where $q$ is the input question, $v_D$ is a document node, $\mathrm{sim}(q,v_D)$ measures the semantic relevance between the question and the document, and $\mathrm{match}(q,v_D)$ measures whether the metadata is consistent with the question, such as company, fiscal year, or sector. The top-ranked documents are kept as the initial search scope.

At the section level, HC-RAG ranks only sections under the selected documents:

\begin{equation}
s_S(q,v_S)=\mathrm{sim}(q,v_S)+\alpha_S \mathrm{prior}(q,v_S),
\end{equation}

where $v_S$ is a section node. The term $\mathrm{prior}(q,v_S)$ represents simple section preference from the question. For example, calculation questions are more likely to require financial statements, while trend or comparison questions may need MD\&A or notes. This step avoids searching all sections in the corpus and keeps the retrieval process closer to the structure of annual reports.

At the evidence-unit level, HC-RAG retrieves text units and table units from the selected sections:

\begin{equation}
s_U(q,v_U)=\mathrm{sim}(q,v_U)+\beta \mathrm{graph}(q,v_U),
\end{equation}

where $v_U$ is a text or table evidence unit. The term $\mathrm{graph}(q,v_U)$ gives extra weight to evidence units that lie on valid document-section-unit paths, match the company and fiscal year in the question, or have useful text-table connections. The final evidence set is selected from the top-ranked evidence units:

\begin{equation}
E_q^*=\mathrm{TopK}(s_U(q,v_U)).
\end{equation}

where $\mathcal{C}_U$ denotes the candidate text and table units from the selected sections.

Compared with flat retrieval, this hierarchical retrieval reduces the search space step by step. Flat retrieval needs to compare the question with a large number of text and table chunks in the whole corpus. HC-RAG first filters documents and sections, then retrieves evidence units only from a smaller graph neighborhood. This makes retrieval more efficient and also keeps clear source paths for the final answer. In our implementation, section nodes are built from the regular HTML structure of Form 10-K filings. For visually complex or scanned filings, the parser can be replaced with layout-aware document models such as LayoutLMv3~\cite{LayoutLMv32022}.

\subsection{Asymmetric Cross-Modal Alignment and Retrieval}
Financial filings usually contain both narrative text and structured tables. Text is often used to explain business reasons, risks, and management views, while tables provide numerical values through row names, column names, reporting periods, units, and cell positions. For example, a question about the reason for a margin decline may depend more on narrative discussion, while a question about calculating the current ratio usually depends on exact values in tables. If text chunks and table units are encoded separately, their similarity scores may not be directly comparable. This can make the retriever miss useful table evidence or mix unrelated textual evidence into the answer.

HC-RAG uses an asymmetric offline-online strategy to reduce this problem. In the offline stage, we use a financial text encoder, such as FinBERT~\cite{Finbert2019}, to encode textual evidence. For table evidence, we use a table-aware encoder, such as TAPAS or TAPEX~\cite{TAPAS2024,TAPEX2022},so that row and column information can be considered. Text units and table units from the same section are treated as related pairs. Their representations are then mapped into the same retrieval space. For a matched text-table pair, we denote their representations as $z^t$ and $z^b$. The basic goal of alignment is to make related text and table evidence closer in the embedding space:

\begin{equation}
\mathcal{L}_{\mathrm{align}} = 1 - \mathrm{sim}(z^t, z^b),
\end{equation}

where $\mathrm{sim}(\cdot,\cdot)$ denotes cosine similarity. In implementation, unrelated text-table pairs in the same batch are also used as negative examples, so the model learns to distinguish matched evidence from unrelated evidence. This follows the common idea of contrastive learning~\cite{oordRepresentationLearningContrastive2018,radfordLearningTransferableVisual2021}, but we keep the formulation simple because the main purpose here is to make text and table evidence comparable during retrieval.

During online inference, HC-RAG does not run a large table encoder for every query. The table encoder is mainly used offline for representation alignment and caching. In the online stage, each table evidence unit is converted into a compact textual form, including its row header, column header, value, and unit. The retriever can then compare the query with both text chunks and flattened table evidence using a single similarity calculation. This design keeps table evidence available during retrieval and avoids the high cost of encoding many table cells for each query. This asymmetric design is a trade-off between accuracy and efficiency. The offline stage uses table-aware models to learn the relation between text and tables. The online stage uses a lighter retrieval form to support faster search over multiple filings. As a result, HC-RAG can retrieve both narrative and tabular evidence without making the inference process too heavy.

\subsection{Intent-Aware Evidence Routing}
\begin{figure*}[t]
    \centering
    \includegraphics[width=0.98\textwidth]{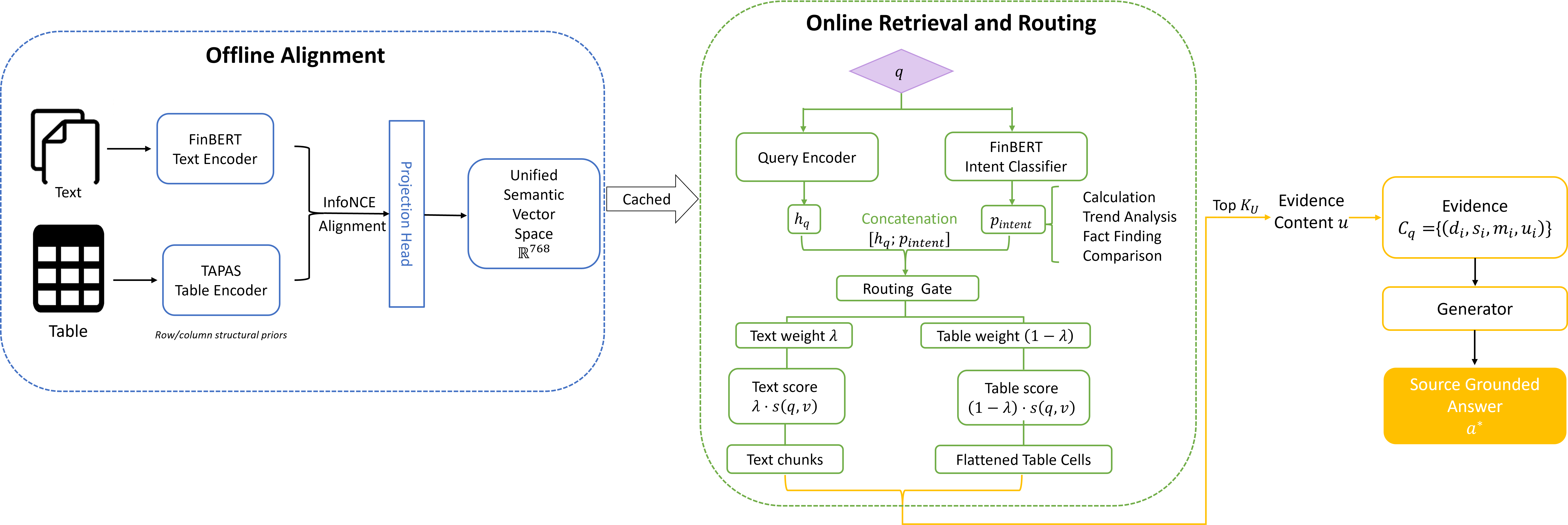}
    \caption{Cross-modal alignment and query-aware routing. In the offline stage, text and table representations are mapped into a shared retrieval space. In the online stage, table evidence is flattened into short strings with row headers, column headers, values, and units. The query encoder and intent classifier produce the query representation and intent distribution, which are used to calculate the routing weight $\lambda$. Text and table evidence are then re-ranked before forming the final evidence set.}
    \label{fig:alignment-routing}
\end{figure*}

Different financial questions need different types of evidence. Calculation questions usually need numerical values from tables. Trend questions need evidence across time. Fact questions often depend on a specific statement or value. Comparison questions need parallel evidence from different companies, metrics, or fiscal years. HC-RAG uses the intent of the query to adjust how text and table evidence are used, as illustrated in the green box in Figure~\ref{fig:alignment-routing}. For each query $q$, we first encode it into a query representation $h_q$. A simple classifier is then used to predict its intent:

\begin{equation}
p_{\mathrm{intent}}=\mathrm{softmax}(W h_q+b),
\end{equation}

where $p_{\mathrm{intent}}$ represents the probability distribution over four intent classes: calculation, trend, fact, and comparison. In our implementation, this classifier can be built on a financial-domain encoder such as FinBERT~\cite{Finbert2019}. Cross-document, cross-year, and hybrid-modal labels are not used as intent labels. They are used later for evidence analysis and evaluation under different data slices. After predicting the query intent, HC-RAG calculates a routing weight $\lambda$:

\begin{equation}
\lambda=\sigma(g(h_q,p_{\mathrm{intent}})),
\end{equation}

where $\lambda$ controls the relative importance of text evidence and table evidence. A larger $\lambda$ means the model gives more weight to textual evidence, while a smaller $\lambda$ means the model gives more weight to tabular evidence. The function $g(\cdot)$ is implemented as a small feed-forward layer. For each candidate evidence unit $v$, its retrieval score is adjusted according to its modality:

\begin{equation}
S(q,v)=
\begin{cases}
\lambda s(q,v), & \text{if } v \text{ is a text unit},\\
(1-\lambda)s(q,v), & \text{if } v \text{ is a table unit}.
\end{cases}
\end{equation}

Here $s(q,v)$ is the original retrieval score between the query and the evidence unit in the shared embedding space. After this re-ranking step, text and table candidates are merged into one evidence list, and the top evidence units are selected for answer generation. In this way, HC-RAG does not always use the same text-table ratio. It changes the evidence preference according to the query intent.

\subsection{Intent-Aware Evidence-Grounded Generation}
The generation layer uses the final retrieved evidence, the predicted intent, and the text-table routing result. Each selected evidence unit is converted into a compact context item:

\begin{equation}
c_i=(d_i,s_i,m_i,u_i),
\end{equation}

$d_i$ is the source filing, $s_i$ is the section, $m_i$ is the modality type, and $u_i$ is the evidence content. Text evidence is kept as short passages. Table evidence is converted into natural language with row headers, column headers, values, and units. The final context is formed by putting these evidence items together:

\begin{equation}
C_q={c_1,c_2,\ldots,c_B}, 
\end{equation}

where $B$ is the evidence budget. HC-RAG then uses different prompting requirements for different intent types. For calculation questions, the prompt asks the model to use exact values, show necessary intermediate steps, and distinguish quoted values from computed results. For trend questions, the prompt asks for the trend direction, related periods, supporting numbers, and possible explanations. For fact questions, the prompt asks for a concise answer with source evidence. For comparison questions, the prompt asks the model to align values across entities or fiscal periods before giving the answer. The final answer is generated as:

\begin{equation}
\hat{a}=\mathrm{LLM}(q,C_q),
\end{equation}

$\hat{a}$ is the generated answer. The generation step is only used after evidence retrieval. HC-RAG first finds the relevant filings, sections, text evidence, and table evidence, then generates the answer based on these selected sources. This also makes the system easier to evaluate at the evidence level, including document hit, section hit, table hit, and cross-document recall.

\section{Proposed Benchmark: Multi-Doc-2025}
In addition to the architecture above, we also introduce \textbf{Multi-Doc-2025}, a benchmark workload for evidence-centric financial QA over SEC Form 10-K filings.\footnote{An anonymized dataset repository is available at \url{https://huggingface.co/datasets/Anonymous-Team-HC-RAG/Multi-Doc-2025}.}It contains 2,327 QA pairs from 179 filings of 87 S\&P 500 companies across 12 GICS sectors and fiscal years 2022--2024. The official split is \textit{primary-company-disjoint}, with 1,600 training, 252 validation, and 475 test examples. No primary company appears in more than one split. For cross-company questions, supporting companies may appear across splits because they serve as comparison evidence rather than the primary query entity. This protocol evaluates generalization to unseen primary companies while preserving realistic peer-comparison workloads.

Table~\ref{tab:dataset_comparison} compares Multi-Doc-2025 with widely used financial QA benchmarks. FinQA and ConvFinQA emphasize numerical reasoning over financial reports~\cite{finqa2021,convfinqa2022}; TAT-QA focuses on table-text reasoning~\cite{zhutatqa2021}; FinanceBench evaluates factual grounding over financial documents~\cite{Financebench}; and DocFinQA extends financial QA to longer documents~\cite{DocFinqa2025}. Multi-Doc-2025 complements these benchmarks by jointly testing cross-document, cross-year, and hybrid-modal reasoning with explicit difficulty tiers.

\begin{table}[H]
\centering
\caption{Comparison with existing financial QA benchmarks. Multi-Doc-2025 jointly evaluates cross-document, cross-year, and hybrid-modal reasoning.}
\label{tab:dataset_comparison}
\setlength{\tabcolsep}{2.6pt}
\renewcommand{\arraystretch}{1.05}
\begin{tabular}{lccccc}
\toprule
\textbf{Benchmark} & \textbf{Scale} & \textbf{Cross-Doc} & \textbf{Cross-Year} & \textbf{Hybrid} & \textbf{Difficulty} \\
\midrule
FinQA~\cite{finqa2021} & 8,281 & No & No & Partial & No \\
TAT-QA~\cite{zhutatqa2021} & 16,552 & No & No & Yes & No \\
FinanceBench~\cite{Financebench} & 150 & Partial & No & Partial & No \\
DocFinQA~\cite{DocFinqa2025} & 7,437 & No & No & Yes & No \\
ConvFinQA~\cite{convfinqa2022} & 3,892 & No & No & Partial & No \\
\textbf{Multi-Doc-2025} & \textbf{2,327} & \textbf{Yes} & \textbf{Yes} & \textbf{Yes} & \textbf{L1--L3} \\
\bottomrule
\end{tabular}
\end{table}

Multi-Doc-2025 is organized into five subsets: single-document fact/calculation, single-document table reasoning, cross-year trend reasoning, cross-company comparison, and full-cross reasoning that combines cross-company, cross-year, and hybrid-modal evidence. The semantic intent space contains four labels: calculation, trend, fact, and comparison. These labels describe the reasoning operation required by the query, while structural attributes such as \texttt{is\_cross\_doc}, \texttt{is\_cross\_year}, and \texttt{is\_hybrid\_modal} describe where the supporting evidence is located. Candidate QA pairs are generated with subset-specific prompts, filtered by rule-based quality checks, and reviewed by a finance domain expert against the original SEC filings. We report EM, token-level F1, numerical execution accuracy, hallucination rate, and slice-wise metrics by intent, subset, and difficulty. Detailed field definitions, examples, subset statistics, and evaluation scripts are provided in the anonymized dataset repository.

\section{Experimental Setup}
\subsection{Datasets}
The paper evaluates HC-RAG on FinQA, TAT-QA, DocFinQA, FinanceBench, and Multi-Doc-2025. These datasets cover numerical reasoning, table-text reasoning, long-document evidence localization, factual grounding, and multi-document aggregation. Multi-Doc-2025 is our proposed benchmark for cross-company, cross-year, and hybrid-modal reasoning; its construction and taxonomy are described in Section 4.

\begin{table}[H]
\centering
\caption{Statistical Information of Experimental Datasets}
\label{tab:dataset_statistics}
\renewcommand{\arraystretch}{1.08}
\setlength{\tabcolsep}{4.5pt}
\begin{tabular}{lccccc}
\toprule
\textbf{Dataset} & \textbf{Samples} & \textbf{Avg. Doc Length} & \textbf{Multi-Doc\%} & \textbf{Table-Related\%} & \textbf{Task Type} \\
\midrule
FinQA          & 8,281  & 512 tokens  & 0\%  & 100\% & Numerical Reasoning \\
TAT-QA         & 16,552 & 480 tokens  & 0\%  & 78\%  & Hybrid Modal QA \\
DocFinQA       & 1,231  & 105k tokens & 0\%  & 100\% & Long-Doc Reasoning \\
FinanceBench   & 150    & 98k tokens  & 12\% & 67\%  & Hallucination Detection \\
Multi-Doc-2025 & 2,327  & 112k tokens & 33.9\% & 49.9\%  & Cross-Doc Reasoning \\
\bottomrule
\end{tabular}
\end{table}

\subsection{Baselines and Fair Comparison Protocol}
Also, we compare HC-RAG with retrieval-based RAG baselines, structured or self-reflective RAG systems, and table-aware or task-specific financial QA models, including BM25-RAG, DPR-RAG~\cite{karpukhinDensePassageRetrieval2020a}, Contriever-RAG, vanilla dense RAG~\cite{RAGNLP,izacardLeveragingPassageRetrieval2020a}, Self-RAG~\cite{asai2024selfrag}, RAPTOR~\cite{sarthi2024raptor}, GraphRAG~\cite{edge2024graphrag}, TAPEX-RAG~\cite{TAPEX2022}, and dataset-specific models when applicable. All RAG-style systems use the same generator, decoding temperature, and maximum context length within each benchmark, with controlled benchmark-specific prompting and evidence serialization; retrieval budgets and top-$k$ cutoffs follow the reported configurations and released experiment logs.

\subsection{Evaluation Metrics}
We report answer-level, evidence-level, and efficiency metrics. Answer quality is measured by EM, token-level F1, Exec-Acc, Hall-Rate, and Faithful Accuracy, with FinanceBench emphasizing faithful accuracy and hallucination rate. For Multi-Doc-2025, we additionally report slice-wise results by semantic intent, subset, difficulty, and evidence-structure attributes. Evidence localization is measured by Doc Hit@5, Section Hit@5, Evidence Recall@5/10, Table Hit@5, and Cross-doc Recall.

\subsection{Implementation Details}
HC-RAG builds the evidence graph from parsed SEC 10-K filings and retrieves evidence through document-, section-, and evidence-unit-level routing. Unless otherwise specified, we set $K_D=5$, $K_S=10$, and $B=20$. The text encoder is initialized from a financial-domain encoder such as FinBERT~\cite{Finbert2019}, table representations are aligned using TAPAS~\cite{TAPAS2024,TAPEX2022}, the intent classifier predicts calculation, trend, fact, and comparison, and all methods use deterministic decoding.

\section{Results and Analysis}
\subsection{Overall Performance Comparison}
Across the five financial QA benchmarks, HC-RAG achieves the strongest answer-level performance, with the largest gains appearing in long-document and multi-document settings. On DocFinQA, HC-RAG improves over RAPTOR by 6.6 F1 points. On Multi-Doc-2025, it reaches 60.2 F1, outperforming GraphRAG and TAPEX-RAG by 10.9 and 6.1 points, respectively (Table~\ref{tab:main_results}). These results suggest that table-aware modeling alone is insufficient for cross-document financial QA; HC-RAG benefits from retrieving evidence along valid document-section-unit paths and routing text-table evidence according to query intent.

\begin{table}[H]
\centering
\caption{Answer-level performance across financial QA benchmarks (\%). Hall Rate is lower better.}
\label{tab:main_results}
\renewcommand{\arraystretch}{1.10}
\setlength{\tabcolsep}{6pt}
\begin{tabular}{lcccccc}
\toprule
\multirow{2}{*}{\textbf{Model}} 
& \multicolumn{1}{c}{\textbf{FinQA}} 
& \multicolumn{1}{c}{\textbf{TAT-QA}} 
& \multicolumn{1}{c}{\textbf{DocFinQA}} 
& \multicolumn{1}{c}{\textbf{FinanceBench}} 
& \multicolumn{1}{c}{\textbf{Multi-Doc-2025}} 
& \multicolumn{1}{c}{\textbf{Hall Rate}$\downarrow$} \\
\cmidrule(lr){2-2}
\cmidrule(lr){3-3}
\cmidrule(lr){4-4}
\cmidrule(lr){5-5}
\cmidrule(lr){6-6}
\cmidrule(lr){7-7}
& EM & Exec Acc & F1 & EM & F1 & \% \\
\midrule
BM25 + DS-V4        & 51.2 & 48.7 & 58.3 & 52.1 & 28.4 & 28.6 \\
DPR + DS-V4         & 56.8 & 54.2 & 63.7 & 58.4 & 34.7 & 25.4 \\
Contriever + DS-V4  & 58.3 & 55.9 & 65.2 & 60.1 & 37.2 & 24.1 \\
Vanilla RAG         & 59.7 & 57.3 & 66.8 & 61.2 & 39.8 & 22.3 \\
Self-RAG            & 62.4 & 60.1 & 69.3 & 64.2 & 44.6 & 19.2 \\
GraphRAG            & 64.1 & 61.8 & 71.5 & 66.3 & 49.3 & 16.8 \\
RAPTOR              & 65.3 & 63.2 & 72.8 & 67.9 & 52.1 & 15.4 \\
TAT-LLM             & 68.7 & 66.4 & 75.2 & 70.1 & 47.8 & 14.1 \\
TAPEX-RAG           & 69.4 & 67.2 & 76.8 & 72.3 & 54.1 & 12.8 \\
\midrule
\textbf{HC-RAG (Ours)} 
                    & \textbf{74.2} 
                    & \textbf{72.8} 
                    & \textbf{79.4} 
                    & \textbf{78.7} 
                    & \textbf{60.2} 
                    & \textbf{11.3} \\
\bottomrule
\end{tabular}
\end{table}

\subsection{Evidence Localization Analysis}
Answer-level gains are meaningful only if the system retrieves the right supporting evidence. We therefore evaluate evidence localization on Multi-Doc-2025 using methods whose retrieved units can be mapped to our evidence schema.

\begin{table}[H]
\centering
\caption{Evidence localization quality on Multi-Doc-2025. Doc Hit and Section Hit measure source localization; Evidence R@$k$ measures fine-grained evidence-unit recall. Table Hit@5 is computed on table-related questions, and Cross-doc Recall is computed on questions requiring multiple filings. ``--'' indicates that the method does not expose comparable table-level evidence identifiers under our evaluation schema.}
\label{tab:evidence_retrieval_results}
\renewcommand{\arraystretch}{1.08}
\setlength{\tabcolsep}{4.5pt}
\begin{tabular}{lcccccc}
\toprule
\textbf{Method} 
& \textbf{Doc Hit@5}$\uparrow$ 
& \textbf{Sec. Hit@5}$\uparrow$ 
& \textbf{Evi. R@5}$\uparrow$ 
& \textbf{Evi. R@10}$\uparrow$ 
& \textbf{Table Hit@5}$\uparrow$ 
& \textbf{Cross-doc R}$\uparrow$ \\
\midrule
BM25-RAG       & 47.16 & 59.37 & 14.39 & 21.89 & --  & 37.52 \\
DPR-RAG        & 46.40 & 57.10 & 15.20 & 22.70 & 12.30 & 35.80 \\
Contriever-RAG & 47.85 & 58.60 & 16.40 & 23.90 & 15.80 & 36.90 \\
Vanilla RAG    & 48.20 & 59.10 & 16.90 & 24.50 & 18.40 & 38.30 \\
GraphRAG       & \textbf{50.10} & 61.80 & 18.70 & 25.10 & -- & 43.20 \\
RAPTOR         & 49.30 & 60.70 & 18.10 & 24.60 & -- & 40.50 \\
TAPEX-RAG      & 47.60 & 60.20 & 19.40 & 25.30 & 66.80 & 39.10 \\
\midrule
\textbf{HC-RAG} 
& 45.26 
& \textbf{64.21} 
& \textbf{22.49} 
& \textbf{26.49} 
& \textbf{89.64} 
& \textbf{48.69} \\
\bottomrule
\end{tabular}
\end{table}

The evidence results show that HC-RAG is not optimized for coarse document hit alone; GraphRAG obtains slightly higher Doc Hit@5, while HC-RAG achieves the strongest section-level localization, fine-grained evidence recall, table retrieval, and cross-document recall (Table~\ref{tab:evidence_retrieval_results}). Compared with the best non-HC-RAG baseline, HC-RAG improves Section Hit@5 to 64.21, raises Table Hit@5 from 66.80 to 89.64, and improves Cross-doc Recall from 43.20 to 48.69. This pattern supports our evidence-centric design: the main advantage of HC-RAG is not merely finding a relevant filing, but navigating to the correct section, table, and cross-document evidence units needed to support the answer.

\subsection{Ablation Study}
We conduct an ablation study on Multi-Doc-2025 to examine the role of each component in HC-RAG. The results are shown in Table~\ref{tab:ablation_results}. Overall, removing the three-level index causes the largest performances drop, which shows that the hierarchical graph structure is important for this task. After removing this index, the model can no longer follow the document-section-unit retrieval process, so it becomes harder to locate evidence in long annual reports. Removing L1 cross-document edges also causes a large decline in Cross-Doc F1, which indicates that questions involving multiple filings need explicit cross-document links. Removing section nodes reduces the overall result as well, suggesting that section-level navigation is useful when the source documents are long and structured.

\begin{table}[H]
\centering
\caption{Ablation Study Results on the Multi-Doc-2025 Dataset}
\label{tab:ablation_results}
\renewcommand{\arraystretch}{1.08}
\setlength{\tabcolsep}{4.2pt}
\begin{tabular}{lccccc}
\toprule
\textbf{Model Variant} 
& \textbf{F1 (\%)} 
& \textbf{EM (\%)} 
& $\boldsymbol{\Delta}$\textbf{F1} 
& \textbf{Cross-Doc F1 (\%)} 
& \textbf{Hybrid Modal F1 (\%)} \\
\midrule
\textbf{HC-RAG (Full)} 
    & \textbf{60.2} 
    & \textbf{53.1} 
    & --- 
    & \textbf{58.7} 
    & \textbf{62.1} \\

w/o Three-Level Index 
    & 49.1 
    & 42.0 
    & -11.1 
    & 38.6 
    & 55.2 \\

w/o Cross-Modal Alignment 
    & 54.3 
    & 47.6 
    & -5.9 
    & 54.8 
    & 49.3 \\

w/o TAPAS 
    & 56.8 
    & 50.4 
    & -3.4 
    & 56.1 
    & 55.7 \\

w/o Query-Aware Fusion 
    & 56.5 
    & 50.0 
    & -3.7 
    & 56.0 
    & 58.2 \\

w/o L1 Cross-Doc Edges 
    & 52.1 
    & 45.7 
    & -8.1 
    & 40.9 
    & 61.0 \\

w/o L2 Section Nodes 
    & 54.9 
    & 48.2 
    & -5.3 
    & 53.0 
    & 56.4 \\

w/o L3 Table Structure 
    & 56.0 
    & 49.1 
    & -4.2 
    & 57.5 
    & 52.0 \\
\bottomrule
\end{tabular}
\end{table}

From the detailed results, the three-level index contributes the most. The full model obtains 60.2 F1, while removing the three-level index reduces F1 to 49.1, with a drop of 11.1 points. This result shows that flat retrieval is not enough for long financial filings. L1 cross-document edges are also important. When these edges are removed, Cross-Doc F1 decreases from 58.7 to 40.9, which means the model loses much of its ability to collect evidence across different filings. Cross-modal alignment mainly affects hybrid-modal questions. Without this component, Hybrid Modal F1 drops from 62.1 to 49.3. Removing TAPAS-based table structure and query-aware fusion also leads to lower results, although the drops are smaller. These results show that table structure and intent-based routing both help the model use textual and tabular evidence more effectively.

\subsection{Efficiency and Scalability Analysis}
HC-RAG trades moderate offline preprocessing for traceable retrieval. Its index construction cost is 4.8 min/doc, lower than GraphRAG's 7.6 min/doc and RAPTOR's 5.3 min/doc, while its inference latency is 3.1 s/query, close to GraphRAG (3.4) and RAPTOR (2.9). This pattern suggests that HC-RAG pays for structured evidence paths once during indexing, then keeps online retrieval bounded by expanding only selected document and section neighborhoods. The full efficiency table and scalability trend are reported in Appendix~\ref{app:additional_analyses} (Table~\ref{tab:efficiency_comparison} and Figure~\ref{fig:scalability}).

\subsection{Adaptive Fusion Analysis}
The learned routing weight $\lambda$ changes with query semantics rather than collapsing to a fixed modality preference. For example, the numerical current-ratio case receives a low $\lambda$ of 0.28 and uses balance-sheet evidence, while the cloud-revenue trend case receives a higher $\lambda$ of 0.71 and relies on MD\&A text. Hybrid and cross-document examples fall between these extremes, such as $\lambda=0.41$ for cross-company R\&D intensity comparison and $\lambda=0.48$ for Intel's margin-decline explanation. Appendix~\ref{app:additional_analyses} provides the full routing distribution and case table (Figure~\ref{fig:fusion-weight-distribution} and Table~\ref{tab:case_analysis}).

\subsection{Robustness Analysis}
HC-RAG is less sensitive to irrelevant distractor evidence because noisy candidates must pass document-level, section-level, and evidence-unit-level filtering before entering the final context. The robustness curve in Appendix~\ref{app:additional_analyses} (Figure~\ref{fig:noise-robustness}) shows the degradation pattern under increasing noise, supporting the role of the hierarchy as a structured noise filter rather than merely a larger retrieval index.

\subsection{Cross-Document Reasoning Diagnostics}
The cross-document diagnostics show that HC-RAG's advantage is strongest when evidence must be aggregated across filings. On the 3+ document setting, HC-RAG reaches 48.7 F1, exceeding GraphRAG by 10.1 points and RAPTOR by 13.5 points. It also leads on YoY EM (57.2), cross-company EM (52.8), and industry-trend F1 (56.4), indicating that Multi-Doc-2025 captures analyst-style reasoning patterns beyond single-filing QA. Appendix~\ref{app:additional_analyses} reports the full diagnostic breakdown in Table~\ref{tab:cross_doc_eval}.

\section{Conclusion}
In this paper, we presented \textbf{HC-RAG}, an evidence retrieval framework for financial question answering over annual reports. Different from methods that split filings into flat chunks, HC-RAG keeps the basic structure of financial reports and retrieves evidence through a document-section-unit process. It first locates relevant filings, then narrows the search to useful sections, and finally retrieves textual and tabular evidence for answer generation. The framework also uses query intent to adjust the use of text and table evidence, so that calculation, trend, fact, and comparison questions can be handled in a more targeted way. Together with \textbf{Multi-Doc-2025}, a benchmark built for cross-company, cross-year, and hybrid-modal financial QA, the experimental results show that HC-RAG improves both answer-level performance and evidence-level localization. The gains are especially clear in section retrieval, table grounding, and cross-document evidence aggregation.

The current framework still has some limitations. HC-RAG relies on the relatively regular structure of SEC-style annual reports, such as stable section titles, HTML filing formats, and available metadata about companies and fiscal years. This makes the method suitable for 10-K filings, but it may not work equally well on scanned reports, noisy PDF files, or documents with irregular layouts. The quality of table extraction can also affect the final result. If a table contains merged cells, missing headers, or unclear units, the flattened table evidence may lose part of the original meaning. Another limitation is that the intent-aware routing module depends on the predicted query intent. When the intent classifier makes mistakes, the model may give too much weight to text or table evidence and retrieve less useful context. In addition, Multi-Doc-2025 mainly focuses on SEC 10-K filings from S\&P 500 companies, so more evaluation is still needed on smaller companies, other financial documents, and non-U.S. filings.

Future work can improve HC-RAG in several directions. One direction is to use more robust document parsing methods, especially for visually complex PDFs and scanned filings. Another direction is to improve table understanding, so that the model can better handle complex headers, units, and multi-level financial tables. It is also useful to make the evidence graph more flexible, allowing it to be updated when new filings are released or when new types of financial evidence are added. Beyond financial QA, the same evidence-centered idea may be applied to other document-heavy domains, such as legal, medical, and regulatory analysis, where answers also need to be supported by clear sources. Overall, the results suggest that domain-specific RAG systems should not only retrieve more context, but should retrieve evidence that is structured, traceable, and directly related to the question.
\clearpage
\newpage
\appendix
\section*{Code and Dataset Availability}
The code and Multi-Doc-2025 benchmark dataset used in this paper are publicly but anonymously available at:
\begin{center}
    \iconlink{https://anonymous.4open.science/r/HC-RAG-Repo-9366/}{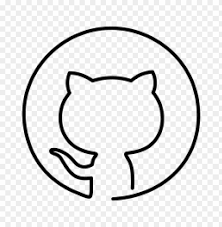}{HC-RAG-Repo}
\quad
\iconlink{https://huggingface.co/datasets/Anonymous-Team-HC-RAG/Multi-Doc-2025}{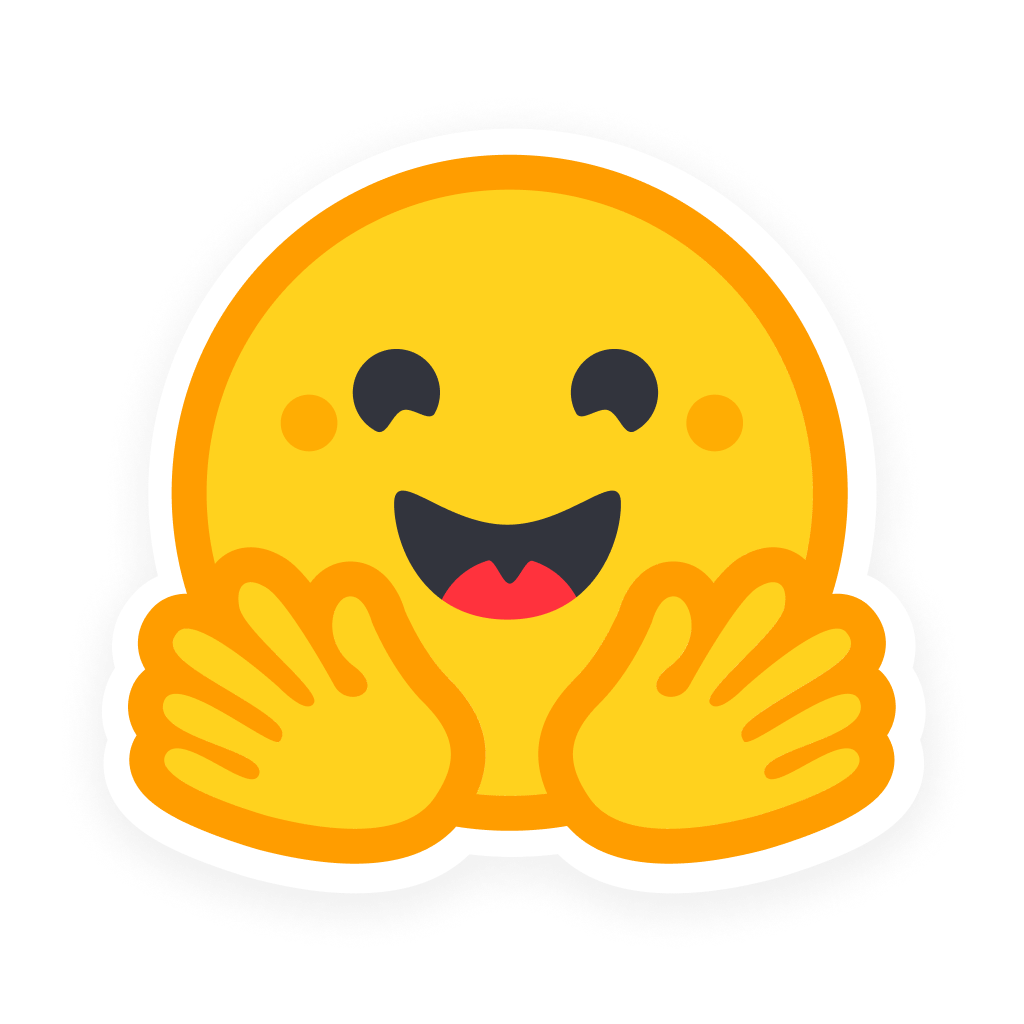}{Multi-Doc-2025}
\end{center}
\section{Additional Experimental Analyses}
\label{app:additional_analyses}

\begin{figure}[H]
    \centering
    \includegraphics[width=0.5\columnwidth]{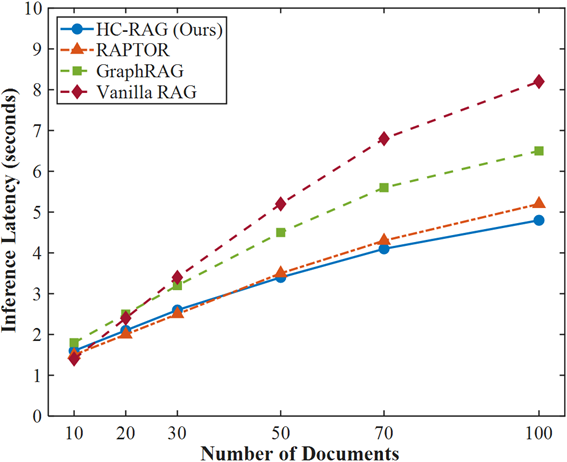}
    \caption{Scalability Analysis}
    \label{fig:scalability}
\end{figure}

\begin{figure}[H]
    \centering
    \includegraphics[width=0.65\columnwidth]{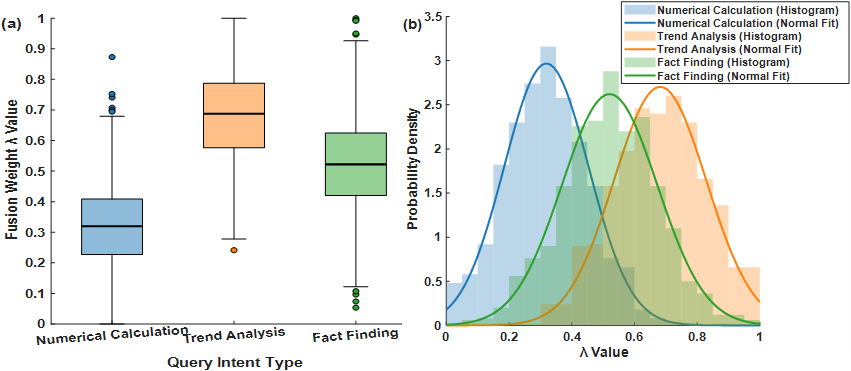}
    \caption{Fusion Weight Distribution Analysis (Composite Figure)}
    \label{fig:fusion-weight-distribution}
\end{figure}

\begin{figure}[H]
    \centering
    \includegraphics[width=0.5\columnwidth]{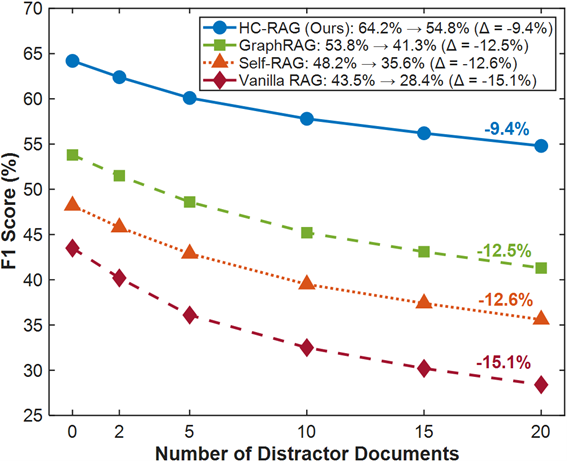}
    \caption{Noise Robustness Experimental Results}
    \label{fig:noise-robustness}
\end{figure}

This appendix complements the main experiments with additional analyses on computational cost, scaling behavior, adaptive routing, robustness, and cross-document reasoning. Table~\ref{tab:efficiency_comparison} and Figure~\ref{fig:scalability} report efficiency and scalability. Figure~\ref{fig:fusion-weight-distribution} and Table~\ref{tab:case_analysis} show the learned routing behavior. Figure~\ref{fig:noise-robustness} and Table~\ref{tab:cross_doc_eval} provide robustness and cross-document diagnostics.

\paragraph{Efficiency and scalability.}
Table~\ref{tab:efficiency_comparison} compares the offline indexing cost, online latency, memory footprint, and model size of HC-RAG against retrieval and reasoning baselines. HC-RAG incurs more indexing overhead than flat dense retrieval because it builds a typed document--section--evidence hierarchy, but its latency remains comparable to graph- and tree-based retrieval systems. The parameter count mainly reflects the combined text and table encoding components used for heterogeneous evidence retrieval.

\begin{table}[H]
\centering
\caption{Efficiency Metrics Comparison}
\label{tab:efficiency_comparison}
\renewcommand{\arraystretch}{1.08}
\setlength{\tabcolsep}{4.5pt}
\begin{tabular}{lcccc}
\toprule
\textbf{Model}
& \textbf{Index Time}
& \textbf{Inference Latency}
& \textbf{GPU Memory}
& \textbf{Parameters} \\
& \textbf{(min/doc)}
& \textbf{(s/query)}
& \textbf{(GB)}
& \textbf{(M)} \\
\midrule
BM25 + DS-V4   & 0.1 & 1.8 & 14.2 & --- \\
DPR + DS-V4    & 1.2 & 2.2 & 18.4 & 110 \\
Vanilla RAG    & 1.2 & 2.3 & 18.4 & 110 \\
Self-RAG       & 1.2 & 3.8 & 21.6 & 110 \\
GraphRAG       & 7.6 & 3.4 & 26.8 & 125 \\
RAPTOR         & 5.3 & 2.9 & 23.2 & 118 \\
TAPEX-RAG      & 5.2 & 2.9 & 27.3 & 516 \\
\textbf{HC-RAG (Ours)}
                & \textbf{4.8}
                & \textbf{3.1}
                & \textbf{24.7}
                & \textbf{517} \\
\bottomrule
\end{tabular}  \\
\footnotesize{
\textbf{Note:} DS-V4 refers to DeepSeek V4-flash.
}
\end{table}

Figure~\ref{fig:scalability} further examines whether retrieval quality and runtime remain stable as the document collection grows. This analysis is intended to isolate the benefit of first narrowing the search space at the document and section levels before ranking fine-grained text and table evidence units.

\paragraph{Adaptive fusion behavior.}
Figure~\ref{fig:fusion-weight-distribution} summarizes the distribution of the routing weight $\lambda$, where larger values assign more weight to textual evidence and smaller values assign more weight to tabular evidence. The distribution provides a global view of whether the fusion module shifts evidence preference across different query types instead of using a fixed text-table mixture.

Table~\ref{tab:case_analysis} gives representative examples of this routing behavior. Numerical questions receive a lower $\lambda$ and are routed toward tabular evidence, while trend-oriented questions receive a higher $\lambda$ and rely more strongly on narrative MD\&A text. Cross-document and hybrid questions fall between these extremes, reflecting their need to combine financial statements with explanatory text.

\begin{table}[H]
\centering
\caption{Case Analysis of Adaptive Fusion}
\label{tab:case_analysis}
\renewcommand{\arraystretch}{1.12}
\setlength{\tabcolsep}{5pt}
\begin{tabular}{p{3.6cm}ccp{2.7cm}p{3.5cm}c}
\toprule
\textbf{Query} 
& \textbf{Query Type} 
& $\boldsymbol{\lambda}$ 
& \textbf{Primary Evidence} 
& \textbf{Answer Summary} 
& \textbf{Correct} \\
\midrule

Calculate Apple’s current ratio for FY2024 
& Numerical 
& 0.28 
& Balance Sheet (Table) 
& Current Ratio = 0.87 
& $\checkmark$ \\

What drove Microsoft’s cloud revenue growth in FY2024 
& Trend 
& 0.71 
& MD\&A Section 7 (Text) 
& Azure growth, enterprise migration... 
& $\checkmark$ \\

When was Alphabet incorporated 
& Fact 
& 0.54 
& Business Overview (Text) 
& October 2, 2015 
& $\checkmark$ \\

Compare R\&D intensity of Apple vs Microsoft in FY2024 
& Cross-Doc 
& 0.41 
& Financial Statements + MD\&A 
& Apple: 7.8\%, Microsoft: 12.3\% 
& $\checkmark$ \\

Explain decline in Intel’s gross margin 
& Hybrid 
& 0.48 
& Income Statement + MD\&A Risk 
& Process costs, competition... 
& $\checkmark$ \\

\bottomrule
\end{tabular}
\end{table}

\paragraph{Robustness and cross-document diagnostics.}
Figure~\ref{fig:noise-robustness} evaluates robustness under additional irrelevant distractor evidence. The comparison focuses on the degradation pattern rather than a single operating point, showing whether hierarchical filtering and evidence-level reranking help preserve answer quality when the retrieved context becomes noisier.

Table~\ref{tab:cross_doc_eval} breaks down performance by multi-document reasoning type. The 2-doc and 3+ doc columns test the number of filings that must be jointly used, while the YoY, cross-company, and industry-trend columns separate common financial comparison patterns. HC-RAG obtains the strongest scores across all listed categories, suggesting that its evidence hierarchy is especially helpful when answer construction depends on aggregating dispersed evidence.

\begin{table}[H]
\centering
\caption{Cross-Document Reasoning Capability Evaluation}
\label{tab:cross_doc_eval}
\renewcommand{\arraystretch}{1.08}
\setlength{\tabcolsep}{4.5pt}
\begin{tabular}{lccccc}
\toprule
\textbf{Model} 
& \textbf{2-Doc F1 (\%)} 
& \textbf{3+ Doc F1 (\%)} 
& \textbf{YoY EM (\%)} 
& \textbf{Cross-Co EM (\%)} 
& \textbf{Industry Trend F1 (\%)} \\
\midrule

Vanilla RAG 
& 38.7 
& 24.3 
& 35.2 
& 28.6 
& 31.4 \\

Self-RAG 
& 45.2 
& 31.8 
& 41.7 
& 35.4 
& 38.2 \\

GraphRAG 
& 52.4 
& 38.6 
& 48.3 
& 42.1 
& 45.7 \\

RAPTOR 
& 49.8 
& 35.2 
& 45.6 
& 38.7 
& 42.3 \\

\textbf{HC-RAG (Ours)} 
& \textbf{61.3} 
& \textbf{48.7} 
& \textbf{57.2} 
& \textbf{52.8} 
& \textbf{56.4} \\

\bottomrule
\end{tabular}
\end{table}


\clearpage
\newpage
\bibliographystyle{unsrtnat}  
\bibliography{references}  

\end{document}